\documentclass[runningheads,a4paper]{llncs}

\usepackage{hyperref}
\usepackage{amssymb}
\usepackage[pdftex]{graphicx}

\begin{document}

\mainmatter  

\title{BetaRun Soccer Simulation League Team:\\ Variety, Complexity, and Learning}

\titlerunning{BetaRun Soccer Simulation League Team}

\author{Olivia Michael \and Oliver Obst}
%
%
%
\institute{
	Centre for Research in Mathematics, Western Sydney University, Locked Bag 1797, Penrith NSW 2751, Australia\\
	EMail: \email{o.obst@westernsydney.edu.au} 
	}


\maketitle


\section{Introduction}

Over the last 20 years, Artificial Intelligence (AI) approaches have demonstrated advances in a number of areas that, originally, were perceived as too hard to tackle by machines at a level of a proficient human. These advances have led to applications of AI in automation, as components of software systems, and are embedded into operating systems of everyday computers. Particular prominent examples, however, resulted from systems that solve difficult benchmark problems to showcase and encourage progress in AI research.
RoboCup offers a set of these benchmark problems in form of official world championships since 1997~\cite{KTS+98}. The most tactical advanced and richest in terms of behavioural complexity of these is the 2D Soccer Simulation League~\cite{PWOJ16}. While the long-standing goal of AI research to beat the human world champion in chess was achieved in the same year of the first RoboCup, a number of more difficult benchmarks and demonstrations were completed in the subsequent 20 years till now. Some examples of such achievements of AI systems are: winning against human experts in Jeopardy~\cite{FBC+10}, learning to play video games from pixel inputs~\cite{MKS+15}, winning against the world champion in the game of Go~\cite{SHM+16}, or beating humans in the game of poker~\cite{BS17}. All of these are impressive results, but differ from RoboCup soccer in that they do not require team work and real-time reactions at the same time. 

The fact that AI systems only just managed to beat human champions in Go or poker could be taken as indication that AI for tasks that require decision making under uncertainty in a team still has a long way to go: Though these benchmarks are very difficult to achieve, they lack some of the complexity of a real-time team task with incomplete information, where opponents can act at any time. Moreover, agents in physical environments like soccer (simulated or real) have an infinite number of possible actions for each agent. 

\section{Towards learning the game of soccer}

Some of the recent achievements of AI systems~\cite{FBC+10,MKS+15,SHM+16,BS17} have in common that they used some form of machine learning as a key component, and that their performance can be immediately compared against the performance of a human player. Teams in the Soccer Simulation play against each other, a comparison against humans is not a part of the competition\footnote{To our knowledge, experimental games in the earlier years of RoboCup against a team of humans using joysticks to control players were easily won by computer programs. It should be noted, however, that the soccer simulation was not designed to be played by humans, and how human players interface with the simulation will obviously affect performance to a large degree.}.
As such, it is impossible to say how well simulation league teams play in terms of potential human performance. It is, however, possible to observe continuing progress in performance of champion teams~\cite{GR11,GFG17}. The small number of teams in the competition (24 qualify for participation in world championships) could also lead to a lack of diversity in behaviours that slows down potential progress.

Even though the complexity of \emph{soccer simulation} may be greater than the complexity of the game Go (due to a practically infinite number of options at every step of the soccer game), the \emph{soccer agents} that participate in RoboCup simulation league are not necessarily more complex (or more ``intelligent'') than the winning Go program \emph{AlphaGo}: instead programmers of a team can choose to reduce complexity by implementing a limited set of tactics that are used as ``canned'' responses to a set of situations, that is, a ``true'' decision will only need to be made at a few critical points, while in most situations players could follow a set script, either explicitly or implicit by the way their options for next actions are constrained by choice. The decision making process of such a team is effectively split over what is programmed into the tactics at the time of creating it, and the run-time selection of an appropriate tactics. The real-time nature of the soccer simulation and the large space of possible actions make it impossible to reason over all options, at every step. 

How exactly these tactics look like, and the way they are developed is naturally very specific to each team, and a number of different paradigms have been used or proposed over the years. A few successful teams employ machine learning techniques to directly learn behaviours, most notably reinforcement learning in, e.g.,~\cite{RM03,GR06,GRT09,GR16}, or planning with MDPs in~\cite{BWC15}. Many other teams use a blend of pre-programmed or reactive scripts, e.g., \cite{PWOJ16,ANH+16,PWO15,PWO14,AN08,MOS02,RLO01}, with a limited use of machine learning. 

So far, in all these cases, use of machine learning has been limited to components or individual skills or behaviours. Almost all teams also  make use of a publicly available base code \emph{agent2D}~\cite{Aki10}, which contains methods to maintain the world model of an agent, as well as a number of (manually implemented) skills. \emph{BetaRun} is a new initiative for a 2D soccer simulation league team, also built on agent2D but otherwise independent from other teams.

\section{Deconstructing World Champions in Go and Robot Soccer}
BetaRun is a new attempt combining both machine learning and manual programming approaches, with the ultimate goal to arrive at a team that is trained entirely from observing and playing games, in the spirit of recent achievements in learning from observations of Atari video games~\cite{MKS+15} or Go~\cite{SHM+16}. As discussed above, this goal is quite ambitious thanks to a dynamic environment that is only partially observable. We are working towards three fundamental changes that have potential for greatly improving performance:
\begin{enumerate}
\item Improving the way the world model is processing and storing information
\item Systematically and automatically improving existing code-base
\item Learning new skills, tactics, and behaviour using reinforcement learning
\end{enumerate}

While we are interested in creating a competitive team, our first focus will be 
to set the foundation for more automatic approaches that only eventually will result in an improvement of performance over the top performing teams. 

\subsection{Improving the way the world model is processing information}

Mobile robots that move around in a known environment together with a team of other robots usually have two very fundamental problems to solve: (1) {\em Where am I?}, and (2) {\em Where is everyone else?}. The current base code agent2D adequately solves these two questions, mostly by triangulation and numerical approaches; while using heuristics to keep track of objects that are not in the current field of view. A further interesting problem for a world model is to answer the question (3) {\em Where will everyone be in the near future?}. 

The self-localisation problem (1) of a robot estimating its own location and orientation, using data from its sensors and known positions of fixed objects can be solved numerically, provided enough of these landmarks are detected by the robot's sensors so that we are able to calculate the current location and orientation. The problem becomes more challenging when perception is limited, and taking into account additional information on previous locations and recent actions of the robot can already help improving estimates of its current pose.  Statistical approaches like Markov Localisation~\cite{FBT99} can be used to reduce some of the uncertainty in estimating and filtering out the most likely current position in the environment.

Robots that cooperate in a team face the additional problem of keeping track of everyone else in the team, in particular with only partial perception of the environment and unreliable information. Again, for objects that are in the field of view, this can be solved numerically once the own position is accurately estimated, and so an accurate model of the current state of the world is beneficial to both plan actions. 

For action planning, it is also beneficial being able to predict what the state of the world might look like in the near future (see also~\cite{GR16}). We are currently investigating learning a representation of current and past sensor data in order to maintain an up-to-date world model, using a recurrent neural network / deep learning based approach~\cite{Sch15}.

\subsection{Systematically and automatically improving existing code-base}

To make use of a more informative world model as described above requires changes to existing skills and behaviours. While it would be an option to manually implement  these improved behaviours, this is not currently our plan for BetaRun. Instead, existing behaviours will be modified to expose all parameters via a shared mechanism, in order to tune performance and maintain their functionality. This can accomplished by a systematical evaluation of performance over sets of games, for a number of opponents~(see, e.g., \cite{PWOJ16}). An alternative approach is to replace existing skills and behaviours of the agent2D base code by ones that are learned by reinforcement learning (see below).  In the intermediate term, the challenge is to arbitrate between behaviours of this code base, and new behaviours as a result of training. 

\subsection{Learning behaviours using reinforcement learning}

The use of reinforcement learning requires substantial changes to both our team code, and also to the tools used around build the team. At the current stage, no component trained by reinforcement learning is part of the code base yet. The plan is to also make use of a predictive world model (see above) for this step of our agenda. For the game of Go, first steps in building \emph{AlphaGo} consisted of training policy networks, i.e., neural networks with weights $\theta$ that predict $p_\theta(a|s)$, the probability of an expert move $a$, in a given situation $s$. This approach does not easily transfer to games with incomplete information, and an infinite number of situations. 
To adopt this idea, each of our players will independently need to make a prediction of a next ``move'', but making this prediction requires completion of some preliminary steps:
\begin{itemize}
\item[--] ``Predict'' the current state of the soccer game: essentially filling in missing information in the world model, as each player perceives only a part of its environment.
\item[--] There is an infinite number of situations on the soccer field\footnote{The number of possible (legal) situations on a $19\times19$ Go board is very large ($> 2 \times10^{170}$), but at least finite~\cite{Tro05}.}, which requires approximations to compare ``situations''. 
\item[--] At every steps of a 6000 cycle soccer simulation, each player can execute elementary actions. These actions are parameterised, practically leading to a large number possible actions. 
Instead of predicting these, predicting higher-level abstract moves that can be achieved by sequences of elementary actions is more useful, but requires (among other things) resolving issues around varying lengths of different such abstract moves, and (ideally) automatically discovering abstract moves. This is an area we are currently investigating.
\end{itemize}

\section{Conclusions}

In BetaRun, we plan to integrate and extend some of the contributions of current machine learning research (e.g.,~\cite{MKS+15,SHM+16}),  ideas from previous competitors~\cite{RM03,GR06,GRT09,GR16,PWO14,PWO15,PWOJ16}. This will result in substantially changing the agent2D framework (or eventually replacing it) that is currently used by most competitors. After 20 years of RoboCup, the Soccer Server~\cite{CDF+03} continues to provide a challenging research problem that is far from ``solved'', while agent2D~\cite{Aki10} provides basic routines and removes some of the barriers to entry into the competition, enabling a diversity of teams that is required for continued progress. Solving some of the current issues in AI research, such as learning winning policies from data for teams from continuous adversarial real-time environments with incomplete information, will require transitioning from teams that make use of the agent2D library to teams that work more directly with the soccer simulator.

\end{document}